\documentclass[10pt]{article}
\usepackage[margin=0.78in]{geometry}
\usepackage{microtype,booktabs,graphicx,float}
\usepackage{amsmath,amssymb,mathtools,bm}
\usepackage{xcolor,enumitem,url}
\usepackage[numbers,sort&compress]{natbib}
\usepackage[colorlinks=true,allcolors=blue!55!black]{hyperref}
\usepackage{tikz}
\usetikzlibrary{arrows.meta,positioning,fit}
\definecolor{brigblue}{HTML}{0072B2}
\definecolor{myopicorange}{HTML}{E69F00}
\newcommand{\method}{BRiG-AFA}
\newcommand{\R}{\mathbb{R}}
\newcommand{\E}{\mathbb{E}}

\title{\textbf{BRiG-AFA: Bellman Risk-to-Go Learning for\\Non-Myopic Active Feature Acquisition}}
\author{Jiaorong Feng\textsuperscript{1,\textdagger} \quad
Qian Li\textsuperscript{2,*} \quad
Ying Li\textsuperscript{2}\\[5pt]
{\small \textsuperscript{1}Curtin Business School, Curtin University,}\\
{\small Perth, Western Australia, Australia}\\[2pt]
{\small \textsuperscript{2}School of Electrical Engineering, Computing and Mathematical Sciences,}\\
{\small Curtin University, Perth, Western Australia, Australia}\\[4pt]
{\small \textsuperscript{\textdagger}Present affiliation: Independent Researcher. \quad
\textsuperscript{*}Corresponding author.}\\[4pt]
{\small \texttt{jiaorong.feng6838@gmail.com} \quad
\texttt{qianli@curtin.edu.au} \quad
\texttt{ying.li@curtin.edu.au}}}
\date{}

\begin{document}
\maketitle

\begin{abstract}
Active feature acquisition (AFA) asks which unobserved feature to measure next for each test instance under a budget. Greedy rules are easy to train but can overlook context features whose value is realized only through later acquisitions, while reinforcement-learning and generative approaches introduce difficult optimization or conditional-density estimation. We introduce \method, a deployable, supervised alternative that learns a separate candidate-conditioned risk-to-go function for every remaining budget. Starting from the one-step terminal classification risk, the functions are fitted backward with Bellman targets; inference greedily minimizes the learned terminal risk using only observed values, the mask, candidate identity, and remaining budget. A controlled non-myopic benchmark shows the expected mechanism: at budgets two and three, \method improves accuracy over its one-step ablation by $4.84\pm2.17$ and $4.39\pm1.10$ percentage points (mean $\pm$ standard error over five seeds). On Fashion-MNIST with 20 candidate pixels, it improves accuracy at every nontrivial reported budget on average, including $10.20\pm0.74$ points at four acquisitions; its mean paired gain across budgets $\{2,4,8,12,16\}$ is $3.50\pm0.37$ points. A three-seed MiniBooNE study is mixed at small budgets but positive at 8 and 16 acquisitions, identifying a current boundary rather than supporting a universal claim. These results establish a reproducible mechanism-level case for direct Bellman risk regression and delimit the experiments still needed for state-of-the-art comparison.
\end{abstract}

\section{Introduction}
Predictive systems often operate before every covariate is known. A clinician may order another test, a sensor may request another channel, or a recognition system may inspect another image region. Because measurements consume money, time, energy, or attention, the relevant decision is not merely which fixed subset is useful, but which feature should be acquired \emph{next}, conditional on values already observed. This is active feature acquisition (AFA), also called dynamic feature selection \citep{shim2018jafa,ma2019eddi,li2021gsm,norcliffe2025sefa}.

The sequential structure matters. An immediately predictive feature is attractive to a myopic policy, yet a weak context feature may reveal which of several downstream measurements is useful. Conditional mutual-information and loss-reduction criteria can therefore miss complementary feature sets \citep{gadgil2024dime,valancius2024aco,norcliffe2025sefa}. Reinforcement learning (RL) can in principle optimize delayed utility, but AFA combines sparse terminal feedback, a changing action set, and a large partially observed state space \citep{shim2018jafa,li2021gsm}. Recent non-greedy methods address this tension with acquisition-conditioned neighbors or stochastic latent encodings \citep{valancius2024aco,norcliffe2025sefa}.

We investigate a simpler question: can non-myopic acquisition be learned by directly regressing the final prediction risk, backward in the remaining budget? \method{} (\textbf{B}ellman \textbf{Ri}sk-to-\textbf{G}o AFA) freezes a predictor trained on partial inputs and fits budget-specific action-value networks. The one-step network predicts the loss after acquiring a candidate. Each longer-horizon network bootstraps from the minimum prediction of the preceding budget network. At deployment, the policy uses no label, hidden regime, or unobserved candidate value.

Our contributions are:
\begin{itemize}[leftmargin=1.4em,itemsep=2pt,topsep=3pt]
    \item a budget-specific Bellman formulation that turns non-myopic AFA into supervised risk regression without an online RL loop or a generative model;
    \item a deployable state representation and generic state-sampling procedure that use no task metadata;
    \item a controlled benchmark that exposes the intended context-then-specialize behavior, plus five-seed Fashion-MNIST evidence and a three-seed MiniBooNE stress test; and
    \item an explicitly paired evaluation against the one-step ablation, separating evidence for non-myopia from gains attributable merely to the predictor.
\end{itemize}

\section{Related Work}
\paragraph{Cost-aware and sequential prediction.}
Early work studied cascades and budgeted prediction, learning when to acquire additional information or invoke a more expensive classifier \citep{trapeznikov2013budget,pu2017acquire}. AFA generalizes this setting to instance-specific feature sequences. JAFA jointly trained a variable-set classifier and a double-Q acquisition agent with stop actions \citep{shim2018jafa}; Opportunistic Learning used deep Q-learning and uncertainty-derived utility for online streams \citep{kachuee2019opportunistic}. Such MDP formulations naturally represent delayed value, but policy optimization can be unstable and sample intensive.

\paragraph{Generative and information-theoretic acquisition.}
EDDI combines a partial VAE with expected information gain about target variables \citep{ma2019eddi}. GSM and GSMRL model arbitrary conditional feature distributions, using them for greedy acquisition or to shape an RL agent \citep{li2021gsm}. These approaches provide counterfactual information about unobserved measurements but require a sufficiently accurate high-dimensional generative model. DIME instead learns discriminative estimates of conditional mutual information, avoiding explicit density estimation while retaining a greedy objective \citep{gadgil2024dime}. GDFS amortizes greedy dynamic feature selection with a scoring network \citep{covert2023gdfs}.

\paragraph{Non-greedy alternatives.}
ACO evaluates candidate subsets using an acquisition-conditioned, nonparametric oracle, explicitly targeting joint informativeness without RL or a deep generative model \citep{valancius2024aco}. SEFA learns stochastic, label-relevant latent encodings and scores acquisitions across possible latent realizations; it was introduced partly to combine non-greedy behavior with supervised training \citep{norcliffe2025sefa}. Explainability-driven ranking more recently distills local feature-importance orders into a decision-transformer policy \citep{guney2025explainability}. \method{} shares the supervised-training motivation but differs in its learning target: it directly approximates the predictor's \emph{terminal risk under a remaining budget}. It neither estimates mutual information nor imputes feature values at inference.

\section{Problem Formulation}
Let $x\in\R^d$ and label $y\in\{1,\ldots,C\}$. A binary mask $m\in\{0,1\}^d$ denotes observed features and $x_m=x\odot m$ is the zero-filled partial input. All experiments here use unit feature costs and a fixed acquisition budget $B$; nonuniform costs are left for future work. A mask-aware predictor
\begin{equation}
 p_\phi(y\mid x_m,m)=\operatorname{softmax} f_\phi(x_m,m)
\end{equation}
is trained before the acquisition policy and then frozen. Given terminal mask $m_B$, the loss is cross-entropy
\begin{equation}
 \ell_\phi(x,y,m_B)=-\log p_\phi(y\mid x\odot m_B,m_B).
\end{equation}
The objective is to learn a deployable policy that minimizes $\E[\ell_\phi]$ after exactly $B$ acquisitions. A valid decision may depend on $x_m$, $m$, and the remaining budget, but not on $y$, unobserved values, or privileged metadata.

\section{BRiG-AFA}\label{sec:method}
\subsection{Budget-specific risk-to-go}
For $r\in\{1,\ldots,B\}$ remaining acquisitions and an available candidate $a$, define $Q_r(x_m,m,a)$ as the conditional expected terminal loss after choosing $a$ and following the learned policy for $r-1$ further acquisitions. Let $e_a$ be the one-hot vector for feature $a$ and let $m^+=m\lor e_a$. The sample-level Bellman targets used for fitted regression are
\begin{align}
 T_1(x,y,m,a)&=\ell_\phi(x,y,m^+),\label{eq:terminal}\\
 T_r(x,y,m,a)&=\min_{a':m^+_{a'}=0}Q_{r-1}(x\odot m^+,m^+,a'),\quad r>1.\label{eq:bellman}
\end{align}
The networks are fitted in ascending order of $r$ with squared error,
\begin{equation}
 \hat\theta_r=\arg\min_{\theta_r}\E_{(x,y),m}\sum_{a:m_a=0}\left(Q_{\theta_r}(x_m,m,a,r/d)-T_r(x,y,m,a)\right)^2.
\end{equation}
Because $Q_{r-1}$ is frozen before $Q_r$ is trained, this is fitted dynamic programming rather than simultaneous temporal-difference learning. Labels are used only to construct training targets. Each Q-network receives $[x_m,m,e_a,r/d]$ and outputs a scalar predicted risk.

For $r>1$, we additionally train on empty-mask actions using targets obtained by greedily rolling out the already fitted shorter-budget networks and evaluating the frozen predictor at the terminal mask. This aligns training with the initial state used at test time. The implementation gives this term unit weight.

\begin{figure}[t]
\centering
\begin{tikzpicture}[font=\small,node distance=5mm and 7mm,box/.style={draw,rounded corners,align=center,minimum height=8mm,inner sep=4pt},arr/.style={-{Latex[length=2mm]},thick}]
\node[box,fill=gray!10] (state) {observed state\\$x\odot m,\;m$};
\node[box,fill=blue!8,right=of state] (cand) {enumerate\\$a\notin m$};
\node[box,fill=blue!13,right=of cand] (q) {$Q_r(x_m,m,a)$\\terminal risk-to-go};
\node[box,fill=green!10,right=of q] (select) {$a^*=\arg\min_a Q_r$};
\node[box,fill=gray!10,right=of select] (reveal) {reveal $x_{a^*}$\\update $m$, $r\!\leftarrow\!r-1$};
\draw[arr] (state)--(cand); \draw[arr] (cand)--(q); \draw[arr] (q)--(select); \draw[arr] (select)--(reveal);
\draw[arr] (reveal.south) -- ++(0,-6mm) -| (state.south);
\node[box,fill=orange!12,below=12mm of q] (train) {offline backward fitting\\$Q_1\leftarrow\ell_\phi$, $Q_r\leftarrow\min Q_{r-1}$};
\draw[arr,dashed] (train)--(q);
\end{tikzpicture}
\caption{\method{} separates offline backward risk fitting from deployable sequential acquisition. At test time only observed values, the mask, candidate identity, and remaining budget enter the policy.}
\label{fig:method}
\end{figure}
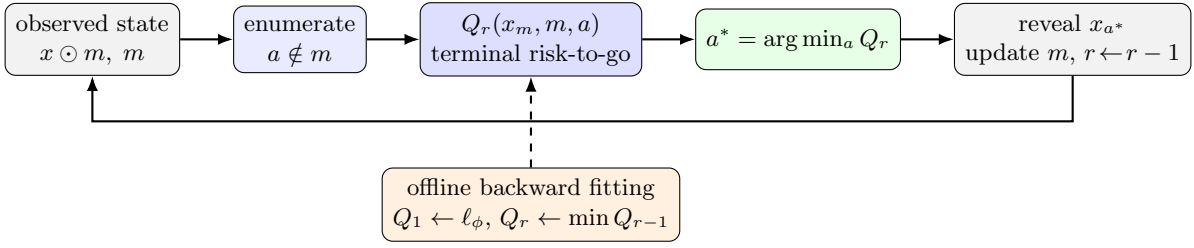

\subsection{State coverage and inference}
For budget level $r$, training masks contain at most $d-r$ observed features. The generic sampler mixes empty masks and randomized feature prefixes of uniformly varying length. It never reads labels, synthetic regimes, or hand-coded feature groups when producing a state. All available candidates are expanded for every sampled state.

At inference, starting with $m=0$, for $t=0,\ldots,B-1$, \method{} selects
\begin{equation}
 a_t=\arg\min_{a:m_a=0}Q_{B-t}(x\odot m,m,a,(B-t)/d),
\end{equation}
reveals only the chosen value, and updates the mask. Our decisive ablation, \emph{myopic Q}, uses $Q_1$ at every step. It has the same predictor, action enumeration, and deployable inputs, isolating the value of the longer-horizon Bellman recursion.

\section{Experimental Design}
\subsection{Datasets and protocol}
\begin{table}[t]
\centering
\caption{Evaluation settings. Reported budgets are test checkpoints; Q-functions are trained for every integer budget up to the largest checkpoint.}
\label{tab:data}
\small
\begin{tabular}{lrrrrl}
\toprule
Dataset & Train & Val. & Test & $d$ & Reported budgets \\
\midrule
CUBE-NM & 6,000 & 2,000 & 2,000 & 12 & 1, 2, 3, 5, 8 \\
Fashion-MNIST-20 & 50,000 & 10,000 & 10,000 & 20 & 1, 2, 4, 8, 12, 16, 20 \\
MiniBooNE & 18,000 & 6,000 & 6,000 & 50 & 1, 2, 4, 8, 16 \\
\bottomrule
\end{tabular}
\end{table}

\paragraph{CUBE-NM.}
We construct a 12-dimensional binary task to require context-dependent acquisition. Feature 0 is a noisy gate identifying one of two regimes; features 1--2 are globally useful; features 3--5 and 6--8 are regime-specific; features 9--11 are distractors. The gate has little direct label value but determines which later group matters. Gaussian label noise has standard deviation 0.25. We use seeds $\{1,3,5,7,9\}$.

\paragraph{Fashion-MNIST-20.}
We use the fixed 20-pixel subset distributed with the SEFA benchmark \citep{norcliffe2025sefa}, retaining the standard 50k/10k/10k split and ten classes. Pixels are scaled to $[0,1]$. The candidate indices are listed in Appendix~\ref{app:details}. We use five seeds $\{1,3,5,7,9\}$.

\paragraph{MiniBooNE.}
We subsample 30,000 examples from the 50-feature binary classification dataset, split 60/20/20, and evaluate seeds $\{1,3,5\}$. This experiment tests whether the mechanism transfers to a larger tabular action space; it is reported as a stress test rather than confirmatory evidence.

\subsection{Models, baselines, and statistics}
The partial-input predictor and each Q-function are two-hidden-layer ReLU MLPs of width 128. Predictor inputs concatenate zero-filled values and masks. Q inputs add a candidate one-hot vector and normalized remaining budget. Both are optimized with Adam at $10^{-3}$; the Q trainer enforces at least eight epochs at each budget level. Global mutual information (MI) ranks features once using the validation split. Random acquisition averages ten independently sampled orders per seed. Myopic Q uses the same learned $Q_1$ at every acquisition. The primary comparison is paired by dataset seed. Curves show mean $\pm$ one standard error; paired-gain panels first subtract within seed and then aggregate, avoiding the inflated uncertainty of unpaired comparisons.

We do not combine results reported in prior papers with our seed-level measurements because their preprocessing, random splits, and acquisition-curve summaries are not identical. The experiments therefore isolate the Bellman-horizon effect through controlled baselines and a matched one-step ablation; comprehensive protocol-matched comparison with recent AFA methods remains future work.

\section{Results}
\subsection{Controlled evidence for non-myopic value}
\begin{figure}[t]
\centering
\includegraphics[width=\linewidth]{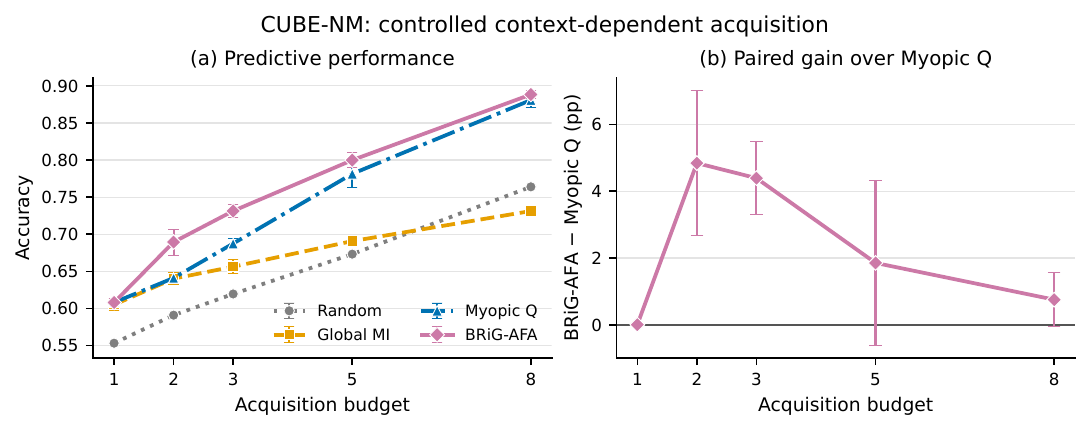}
\caption{CUBE-NM results over five paired seeds. Left: accuracy by acquisition budget. Right: within-seed accuracy gain of \method{} over myopic Q. The gain peaks where acquiring the weak context feature can redirect later acquisitions. Error bars show one standard error.}
\label{fig:cube}
\end{figure}

On CUBE-NM, the two Q policies are identical at one acquisition by construction. At budget two, \method{} reaches 68.95\% accuracy compared with 64.11\% for myopic Q, a paired gain of $4.84\pm2.17$ points. At budget three, it reaches 73.14\% versus 68.75\%, a $4.39\pm1.10$ point gain that is positive in all five seeds. Gains contract to $1.85\pm2.46$ and $0.75\pm0.81$ points at budgets five and eight. This profile matches the proposed mechanism: foresight matters most when the budget is sufficient to act on context but still restrictive.

\subsection{Fashion-MNIST}
\begin{figure}[t]
\centering
\includegraphics[width=\linewidth]{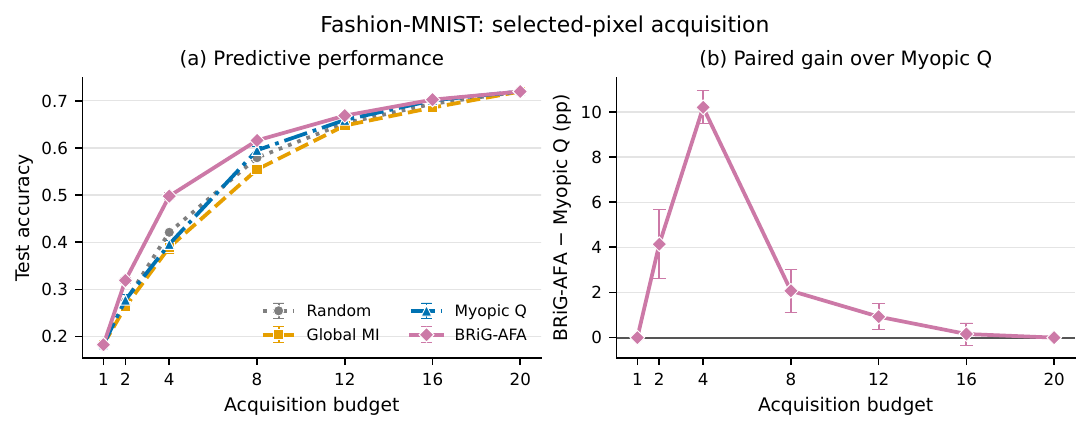}
\caption{Fashion-MNIST with 20 candidate pixels, five paired seeds. \method{} provides its largest advantage at four acquisitions and converges with alternatives as most candidates become observable.}
\label{fig:fashion}
\end{figure}

\begin{table}[t]
\centering
\caption{Paired \method{} minus myopic-Q accuracy gains (percentage points). ``Positive seeds'' counts seed-level improvements.}
\label{tab:gains}
\small
\begin{tabular}{lrrrr}
\toprule
Dataset & Budget & Mean gain & Paired SE & Positive seeds \\
\midrule
CUBE-NM & 2 & 4.84 & 2.17 & 4/5 \\
CUBE-NM & 3 & 4.39 & 1.10 & 5/5 \\
Fashion-MNIST-20 & 2 & 4.14 & 1.53 & 4/5 \\
Fashion-MNIST-20 & 4 & \textbf{10.20} & 0.74 & 5/5 \\
Fashion-MNIST-20 & 8 & 2.07 & 0.96 & 4/5 \\
Fashion-MNIST-20 & 12 & 0.93 & 0.57 & 4/5 \\
Fashion-MNIST-20 & 16 & 0.16 & 0.49 & 3/5 \\
MiniBooNE & 8 & 0.93 & 1.33 & 2/3 \\
MiniBooNE & 16 & 0.93 & 0.27 & 3/3 \\
\bottomrule
\end{tabular}
\end{table}

Fashion-MNIST provides the strongest real-data result. At four acquisitions, \method{} obtains 49.78\% accuracy versus 39.58\% for myopic Q, producing a $10.20\pm0.74$ point paired gain that is positive for every seed. Across the nontrivial budgets $\{2,4,8,12,16\}$, mean accuracy is 56.07\% for \method{}, 52.57\% for myopic Q, 50.80\% for global MI, and 52.40\% for random acquisition. The per-seed difference averaged across these budgets is positive for all five seeds, with mean $3.50\pm0.37$ points (paired standard error; approximate $t_4$ 95\% interval $[2.47,4.53]$). As the budget approaches all 20 pixels, the policies necessarily converge.

\begin{figure}[t]
\centering
\includegraphics[width=0.94\linewidth]{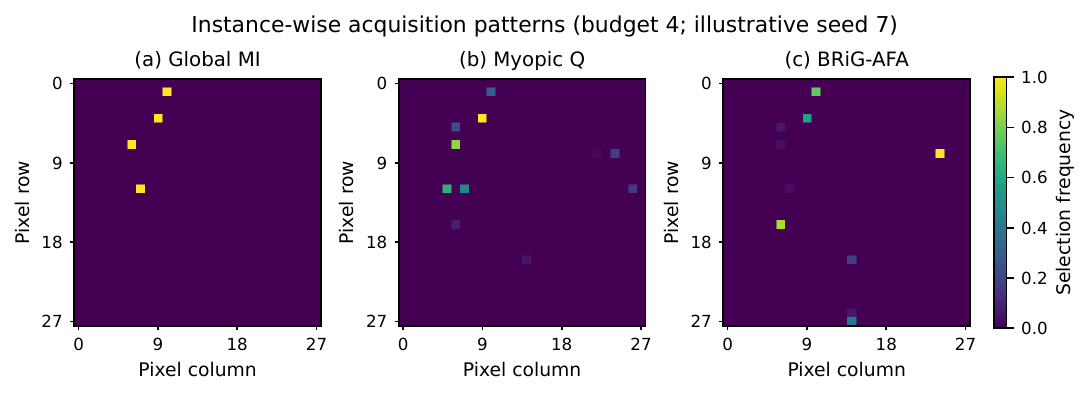}
\caption{Illustrative Fashion-MNIST acquisition frequencies at budget four (seed 7). Each panel maps the 20 candidate coordinates back to the $28\times28$ image. BRiG's distribution differs from both the static global order and the repeatedly applied one-step rule. This is descriptive evidence, not a significance test.}
\label{fig:maps}
\end{figure}

The acquisition maps in Figure~\ref{fig:maps} provide a qualitative sanity check. Global MI concentrates on its fixed top-ranked coordinates, whereas the learned policies distribute actions across instance-dependent locations. BRiG and myopic Q also differ despite sharing $Q_1$, consistent with budget-conditioned planning changing early actions.

\subsection{Where the advantage does not uniformly transfer}
On MiniBooNE, \method{} is worse than myopic Q at budgets two and four ($-0.78\pm0.73$ and $-0.98\pm0.36$ points) but better at eight and sixteen ($0.93\pm1.33$ and $0.93\pm0.27$ points). The final gain is positive in all three seeds. This mixed curve is important: bootstrapped risk regression can accumulate approximation error, and longer horizons are not automatically useful when the one-step ranking already captures much of the signal. We therefore treat CUBE-NM as mechanism validation, Fashion-MNIST as the principal empirical result, and MiniBooNE as a boundary case.

\section{Discussion and Limitations}
The experiments support a focused conclusion: with the predictor, state sampler, and evaluation protocol held fixed, budget-specific risk-to-go can outperform repeatedly applying a one-step action value. They do not establish broad superiority over modern AFA systems. A broader empirical assessment requires SEFA, ACO, DIME, and GDFS to be evaluated under matched splits, preprocessing, candidate sets, budget integration, and repeated seeds.

Several technical limitations remain. First, the Bellman targets bootstrap through learned networks and may compound error as $r$ increases. Second, enumerating every available action costs $O(Bd)$ Q evaluations per instance without batching or candidate pruning. Third, the current predictor is trained with randomly sampled masks rather than jointly with the induced acquisition distribution. Fourth, all experiments use unit costs and fixed horizons; stop actions and heterogeneous measurement costs require a cost-aware Bellman formulation. Fifth, three datasets---one synthetic---are insufficient for claims across modalities or missingness mechanisms. Finally, all training data are fully observed. In retrospective domains, policy-induced distribution shift and the validity of offline evaluation require separate causal assumptions and estimators \citep{vonkleist2023evaluation}.

\section{Conclusion}
We presented \method, a supervised AFA method that learns terminal classification risk backward over the remaining acquisition budget. The controlled benchmark and Fashion-MNIST results show that the budget-specific policy can exploit delayed value that its one-step ablation misses, while MiniBooNE reveals that the benefit is not universal. The central practical lesson is that direct, deployable Bellman risk regression is a viable middle ground between greedy utility estimation and full RL or generative planning. The next step is not a broader claim, but a protocol-matched benchmark against modern non-greedy AFA methods and extensions to costs and adaptive stopping.

\section*{Code Availability}
The source code are available at \url{https://github.com/JIAORONG-FENG/BRiG-AFA/tree/main}.

\appendix
\section{Implementation and Reproducibility Details}\label{app:details}
\noindent\textbf{Fashion-MNIST candidates.} In flattened row-major indexing, the 20 pixels are {\small 10, 38, 121, 146, 202, 246, 248, 341, 343, 362, 406, 434, 454, 490, 546, 574, 580, 602, 742, 770.}

\noindent\textbf{Training configuration.} For a maximum budget $B$, we fit $Q_1,\ldots,Q_B$ consecutively, including intermediate budgets not reported at evaluation. The frozen predictor and each Q-function are two-layer, width-128 ReLU MLPs. A Q-function receives a $3d+1$ dimensional input. Training uses cross-entropy terminal risk, generic sampled states, and the empty-mask rollout term described in Section~\ref{sec:method}.

\section{Cross-Entropy Curves}
\begin{figure}[H]
\centering
\includegraphics[width=0.90\linewidth]{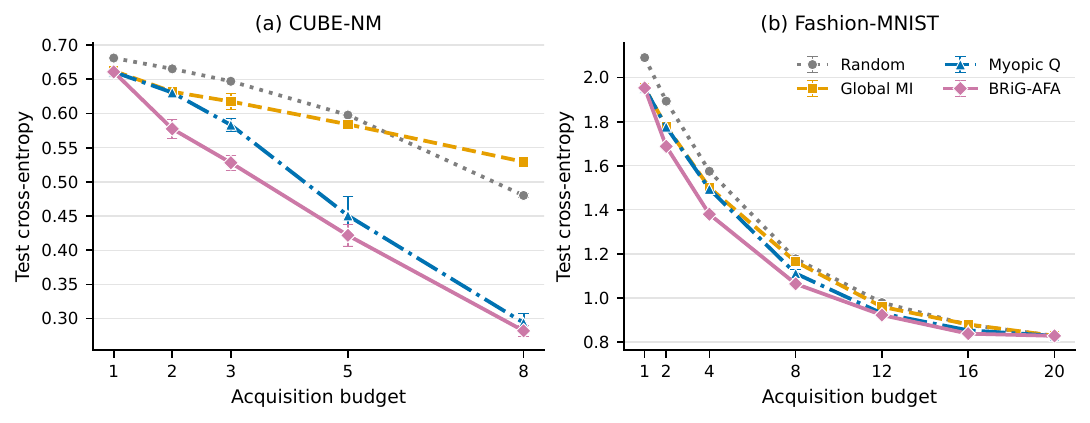}
\caption{Terminal cross-entropy for CUBE-NM and Fashion-MNIST. Lower is better; ribbons/error bars denote one standard error over seeds.}
\label{fig:ce}
\end{figure}

\section{MiniBooNE Stress Test}
\begin{figure}[H]
\centering
\includegraphics[width=0.90\linewidth]{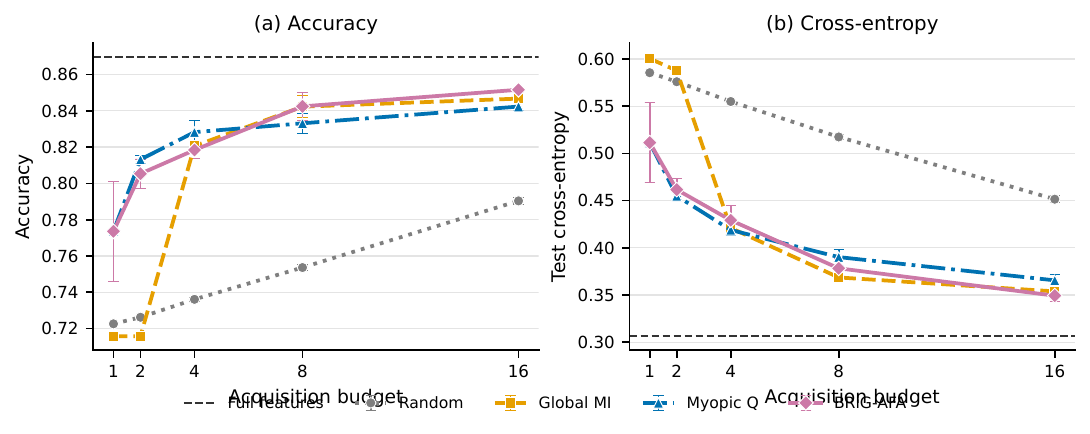}
\caption{MiniBooNE accuracy and terminal cross-entropy over three seeds. BRiG-AFA trails myopic Q at budgets 2 and 4 but exceeds it at budgets 8 and 16 in accuracy. Error bars denote one standard error.}
\label{fig:miniboone}
\end{figure}

\section{CUBE-NM Diagnostic References}
\begin{figure}[H]
\centering
\includegraphics[width=0.56\linewidth]{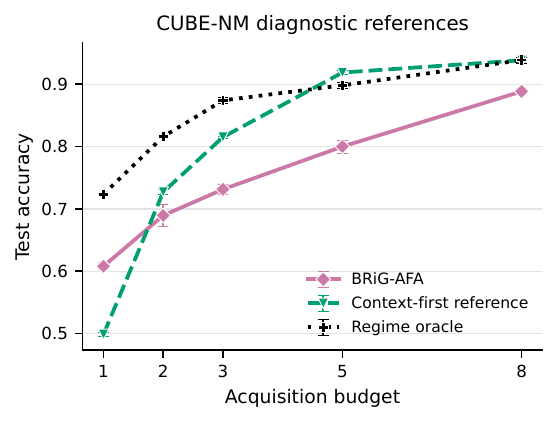}
\caption{Diagnostic CUBE-NM policies. Context-first uses known task structure, whereas regime oracle receives the true regime at no acquisition cost. These policies are not deployable and are excluded from the main comparison.}
\label{fig:references}
\end{figure}
\clearpage

\bibliographystyle{plainnat}
\bibliography{references}
\end{document}